\documentclass[runningheads]{llncs}
\usepackage[T1]{fontenc}
\usepackage{multirow} 
\usepackage{booktabs} 
\usepackage{xcolor}
\usepackage{colortbl}
\usepackage{graphicx,verbatim}
\usepackage{amsmath} 
\usepackage{amssymb}
\usepackage{subcaption}
\usepackage[colorlinks=true,citecolor=green]{hyperref}
\usepackage[misc]{ifsym}
\newcommand{\para}[1]{\vspace{.05in}\noindent\textbf{#1}}

\begin{document}
%
\title{SegDINO: Introducing Multi-Scale Structure into DINO for Efficient Medical Image Segmentation}
\titlerunning{SegDINO: Multi-Scale Adaptation of DINOv3 for Medical Segmentation}

\author{
Sicheng Yang\inst{1} \and
Hongqiu Wang\inst{1} \and
Zhaohu Xing\inst{1} \and
Sixiang Chen\inst{1} \and
Qiuxia Yang\inst{2} \and
Yize Mao\inst{3} \and
Guang Yang\inst{4} \and
Lei Zhu\inst{1}\textsuperscript{\Letter}
}

\authorrunning{S. Yang et al.}

\institute{
The Hong Kong University of Science and Technology (Guangzhou), Guangzhou, China
\and
Department of Radiology, State Key Laboratory of Oncology in South China, Guangdong Provincial Clinical Research Center for Cancer, Sun Yat-sen University Cancer Center, Guangzhou 510060, China
\and
Department of Pancreatobiliary Surgery, State Key Laboratory of Oncology in South China, Sun Yat-sen University Cancer Center, Guangzhou, Guangdong, China
\and
Imperial College London, London, United Kingdom
}
  
\maketitle              
\begingroup
\renewcommand{\thefootnote}{\Letter}
\footnotetext{Lei Zhu (leizhu@hkust-gz.edu.cn) is the corresponding author.}
\endgroup
\begin{abstract}
Self-supervised DINO models provide strong transferable visual representations, yet applying them directly to image segmentation remains challenging. Existing approaches commonly rely on heavy decoders with complex upsampling, introducing substantial parameter and computational overhead. We observe that introducing scale into DINO features is far more critical than increasing decoder capacity. In this work, we present SegDINO, an efficient segmentation framework that integrates a DINOv3 backbone with lightweight scale modeling. SegDINO introduces Token Pyramid Adaptation (TPA) to reorganize intermediate DINO features into a pseudo multi-scale hierarchy, and Scale-Aware Decoding (SAD) for efficient intra-scale refinement and top-down multi-scale propagation. We further curate PanCT, a new CT dataset containing 284 patients with expert-annotated pancreatic tumors, to assess SegDINO's ability to handle difficult small-lesion cases. Extensive experiments on PanCT and three public benchmarks demonstrate that SegDINO achieves state-of-the-art results with high efficiency. The code is available at \url{https://github.com/script-Yang/segdino_v2}.

\keywords{Medical Image Segmentation  \and Multi-Scale Representation }

\end{abstract}
\section{Introduction}
Medical image segmentation plays a central role in medical image analysis, serving as a foundation for downstream clinical applications, such as treatment planning and computer-aided diagnosis~\cite{azad2024medical,wang2024dual,wang2024video}.
Despite remarkable progress achieved by convolutional networks~\cite{ronneberger2015u}, transformer-based models~\cite{li2024transformer}, diffusion-based architectures~\cite{wu2024medsegdiff}, and Mamba-based frameworks~\cite{liu2024swin,wang2025serp}, these approaches often struggle to achieve strong generalization in medical imaging scenarios, particularly when training data are limited or acquired under varying imaging protocols~\cite{zhang2021understanding}.
Recent SAM-based segmentation models~\cite{kirillov2023segment,xiong2026sam2} exhibit strong zero-shot abilities, but their heavy architectural designs result in high fine-tuning costs when adapted to downstream tasks, limiting their practical efficiency.

Rather than training segmentation networks entirely from scratch, a growing line of work instead leverages pretrained visual representations to obtain richer semantic and structural priors~\cite{zhou2024image}.
Within this paradigm, self-supervised foundation models from the DINO family have emerged as especially powerful due to their strong cross-domain transferability~\cite{wang2025dinov3,yang2026vaevq,yang2026vq}. DINO~\cite{caron2021emerging} and DINOv2~\cite{oquab2023dinov2} have proven effective in general-purpose representation learning~\cite{zhu2024scaling}, enabling robust performance on downstream detection~\cite{damm2025anomalydino} and segmentation tasks~\cite{ayzenberg2024dinov2,gao2025dino,li2025meddinov3}. Most recently, DINOv3~\cite{simeoni2025dinov3} introduced substantial improvements in self-supervised pretraining, providing even stronger invariance and scalability.

However, effectively adapting DINO-based representations to segmentation tasks remains a non-trivial challenge. In practice, we observe that features extracted by DINO lack an explicitly defined multi-scale hierarchy, which makes it difficult for overly simple decoders to capture fine-grained semantic structures, particularly for subtle targets (e.g., small lesion segmentation). Several prior works adopt relatively heavy decoder designs, such as stacking multiple multi-scale fusion modules~\cite{gao2025dino} or employing complex upsampling pipelines~\cite{yang2025unimatch}. These designs largely inherit architectural principles from semantic segmentation networks originally developed for natural images, and while effective, they introduce substantial parameter overhead and computational cost. We argue that, given the strong representational capacity of DINO features, the key challenge lies in carefully designing scale modeling mechanisms, rather than increasing decoder capacity. With an appropriately designed decoding strategy, competitive segmentation performance can be achieved with significantly fewer parameters.

In this paper, we introduce SegDINO, an efficient segmentation framework that adapts DINO-based representations for medical image segmentation using scale modeling. We design Token Pyramid Adaptation (TPA) to reorganize representations extracted from different DINO depths into a pseudo pyramid, injecting scale diversity into patch-level features. We further introduce Scale-Aware Decoding (SAD), a lightweight decoding strategy that combines intra-scale refinement with top-down inter-scale propagation.
Finally, we propose a new dataset for small-lesion medical lesion segmentation, named PanCT, which includes 284 patients collected from the Radiology Department, with primary lesions annotated by two experienced radiologists.
Extensive experiments on four datasets demonstrate that SegDINO consistently achieves superior segmentation performance while maintaining high efficiency.
\section{Method}
\begin{figure*}[t]
\centering
\includegraphics[width=0.99\linewidth]{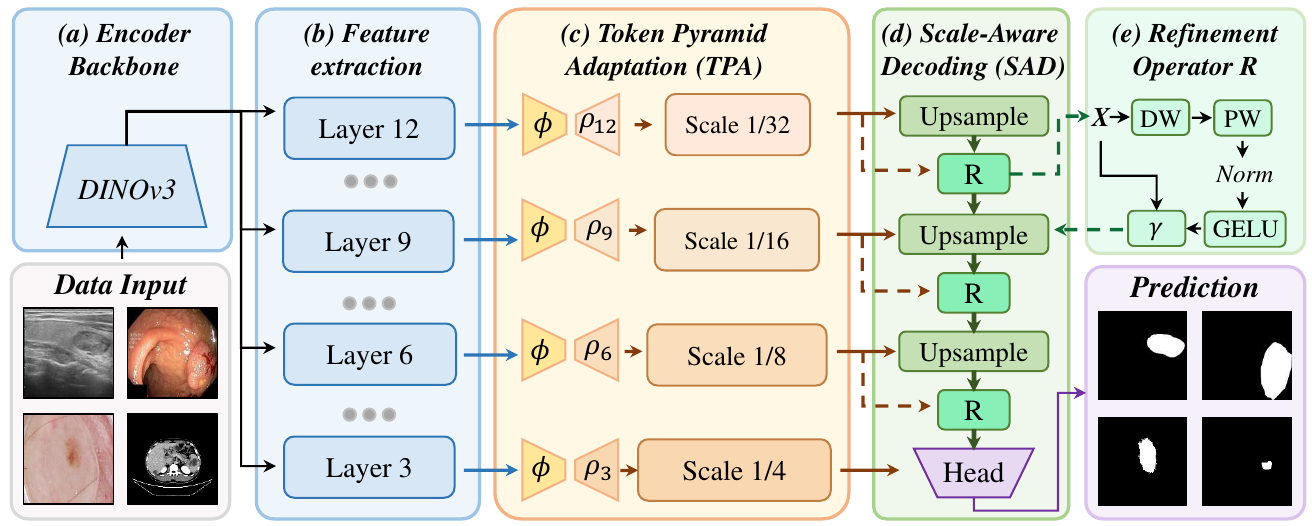}
\caption{An overview of the proposed SegDINO framework. A frozen DINOv3 encoder extracts intermediate features from multiple layers. These features are first projected and reorganized by the Token Pyramid Adaptation (TPA) module to construct a pseudo multi-scale pyramid, introducing scale-aware representations. The resulting pyramid is then processed by the Scale-Aware Decoder (SAD), which performs intra-scale refinement and inter-scale information propagation using a residual refinement operator $R$. Finally, a lightweight prediction head produces the segmentation output.}
\label{fig:overview}
\end{figure*}

SegDINO mainly consists of three components: (1) a DINO-based feature extraction module that collects intermediate representations from multiple depths to capture both low-level structures and high-level semantics, (2) a Token Pyramid Adaptation (TPA) module that reorganizes these representations into a pseudo pyramid by introducing scale diversity, and (3) a Scale-Aware Decoding (SAD) module that performs lightweight intra-scale refinement and inter-scale propagation to generate the final segmentation prediction. Fig.~\ref{fig:overview} illustrates the overall architecture of SegDINO. 
We describe the details in the following subsections.

\subsection{Encoder backbone}
We adopt a pretrained DINOv3~\cite{simeoni2025dinov3} as the encoder. As illustrated in Fig.~\ref{fig:overview} (a), given an input image $\mathbf{x}\in\mathbb{R}^{H\times W\times 3}$, it is partitioned into $N=(H/p)\times(W/p)$ non-overlapping patches and embedded into an initial token sequence $\mathbf{Z}^{(0)}\in\mathbb{R}^{N\times d}$.
The tokens are then iteratively updated by a stack of $L$ Transformer blocks $\mathcal{B}_\ell$ as
\begin{align}
\mathbf{Z}^{(\ell)}=\mathcal{B}_\ell\! \ \big(\mathbf{Z}^{(\ell-1)}\big),\qquad \ell=1,\dots,L.
\end{align}
As shown in Fig.~\ref{fig:overview} (b), to harvest both low-level structure and high-level semantics, we collect intermediate token matrices from a subset of layers
\begin{align}
\mathcal{L}=\{\ell_1,\ell_2,\dots,\ell_K\}\subseteq\{1,\dots,L\}.
\end{align}
For each $\ell_k\in\mathcal{L}$, we directly take the patch tokens $\mathbf{Z}_{p}^{(\ell_k)}\in\mathbb{R}^{N\times d}$ from the ViT output and discard any non-patch tokens (e.g., class or register tokens). 
The encoder output is the multi-level token set
$\big\{\mathbf{Z}_{p}^{(\ell_k)}\big\}_{k=1}^{K}.
$

\subsection{Token Pyramid Adaptation}

The intermediate token set $\{\mathbf{Z}_{p}^{(\ell_k)}\}_{k=1}^{K}$ extracted from the DINO encoder captures representations of increasing semantic abstraction across transformer depth.
However, all token matrices are defined on the same patch grid and therefore share an identical spatial resolution.
Such a uniform-scale representation is suboptimal for image segmentation, where hierarchical feature scales are essential for integrating global semantic context with fine-grained boundary details.

As illustrated in Fig.~\ref{fig:overview}(c), we introduce a simple yet effective \emph{Token Pyramid Adaptation (TPA)} strategy to explicitly inject scale diversity into DINO features. 
Inspired by the feature pyramid design in FPN~\cite{lin2017feature}, TPA reorganizes intermediate DINO tokens into a hierarchy of spatial feature maps. 
Specifically, for each intermediate token matrix $\mathbf{Z}_{p}^{(\ell_k)}\in\mathbb{R}^{N\times d}$, we reshape the patch tokens back to a 2D feature map according to the original patch layout, thereby restoring their spatial arrangement.
\begin{align}
\mathbf{F}_k = \mathrm{reshape}\!\left(\mathbf{Z}_{p}^{(\ell_k)}\right)\in\mathbb{R}^{d\times H_p\times W_p},
\end{align}
where $H_p=H/p$ and $W_p=W/p$ denote the height and width of the patch grid.

Since different transformer layers may produce features with heterogeneous channel statistics, we apply a $1\times1$ convolution to project all feature maps into a unified embedding space:
\begin{align}
\tilde{\mathbf{F}}_k = \phi(\mathbf{F}_k), \qquad \phi:\mathbb{R}^{d}\rightarrow\mathbb{R}^{d'}.
\end{align}
This projection preserves spatial correspondence while enabling subsequent cross-level interactions. Next, to introduce spatial hierarchy for dense decoding, we design scale-specific spatial transformations to the projected features:
\begin{align}
\mathbf{P}_k = \rho_k(\tilde{\mathbf{F}}_k),
\end{align}
where $\rho_k(\cdot)$ denotes a lightweight resizing operator, implemented using strided convolution.
As a result, TPA constructs a pseudo feature pyramid with progressively varying spatial resolutions (e.g., $\tfrac{1}{4}$, $\tfrac{1}{8}$, $\tfrac{1}{16}$, and $\tfrac{1}{32}$ of the input resolution), while preserving the hierarchical semantic representations extracted by DINO.

\subsection{Scale-Aware Decoding}

With the pseudo pyramid $\mathcal{P}=\{\mathbf{P}_k\}_{k=1}^{K}$ constructed by the TPA, as shown in Fig.~\ref{fig:overview} (d), we introduce a \emph{Scale-Aware Decoding (SAD)} strategy to generate the final segmentation prediction in a lightweight manner. The decoding process is decomposed into two complementary components: \emph{intra-scale updating}, which performs preliminary refinement within each individual scale, and \emph{inter-scale propagation}, which enables information transfer across different spatial resolutions. This design allows the decoder to progressively integrate coarse semantic information with fine-grained spatial details while avoiding heavy modules.

\para{Lightweight residual refinement.}
As illustrated in Fig.~\ref{fig:overview} (e), we first define a lightweight residual refinement operator $\mathcal{R}(\cdot)$ to update features:
\begin{align}
\mathcal{R}(\mathbf{X}) = \mathbf{X} + \gamma \cdot \sigma\!\left(\mathrm{Norm}\!\left(\mathrm{PW}\!\left(\mathrm{DW}(\mathbf{X})\right)\right)\right),
\end{align}
where $\mathrm{DW}(\cdot)$ and $\mathrm{PW}(\cdot)$ denote depthwise and pointwise convolutions, respectively, $\mathrm{Norm}(\cdot)$ is GroupNorm, $\sigma(\cdot)$ is the GELU activation, and $\gamma$ is a learnable residual scaling parameter initialized to $0$ for stable optimization. This operator introduces only minimal computational overhead.

\para{Intra-scale updating.}
After spatial resizing and channel unification in TPA, each pyramid level is first pre-conditioned independently:
\begin{align}
\mathbf{L}_k = \mathcal{R}\!\left(\mathbf{P}_k\right), \qquad k=1,\dots,K .
\end{align}
This step constructs scale-specific representations by locally refining features within each resolution, preparing them for subsequent multi-scale interaction.

\para{Inter-scale propagation.}
To fuse representations across different spatial scales, we employ a top-down inter-scale pathway that progressively integrates coarse- and fine-scale features.
Let $\mathrm{Up}(\cdot;\mathbf{L}_k)$ denote bilinear interpolation that resizes its input to match the spatial resolution of $\mathbf{L}_k$. Starting from the coarsest level, inter-scale propagation is performed recursively:
\begin{align}
\mathbf{X}_K &= \mathcal{R}\!\left(\mathbf{L}_K\right), \\
\mathbf{X}_k &= \mathcal{R}\!\left(\mathrm{Up}(\mathbf{X}_{k+1};\mathbf{L}_k) + \mathbf{L}_k\right), \qquad k=K-1,\dots,1 .
\end{align}

Finally, the segmentation prediction is obtained from the finest-scale representation using a linear prediction head:
\begin{align}
\mathbf{Y} = \mathcal{H}(\mathbf{X}_1),
\end{align}
where $\mathcal{H}(\cdot)$ is implemented as a $1\times1$ convolution. Through the proposed Scale-Aware Decoding, multi-scale features extracted by TPA are efficiently decoded into the final segmentation output.
\section{Experiments}
\subsection{Datasets}
\para{PanCT Dataset.}
We collected PanCT from the Radiology Department, comprising 284 patients with confirmed pancreatic cancer, including 243 patients for training and 41 for internal testing. 
The training cohort has a median age of 59 years with 94 female and 149 male patients, while the test cohort has a median age of 56 years with 20 female and 21 male patients. 
All scans were independently annotated by two experienced radiologists, and all patient data were de-identified under institutional ethical approval. 
Each 3D CT volume is converted into 2D axial slices for efficient slice-level modeling and fair comparison with 2D baselines; we acknowledge the loss of inter-slice context and leave 2.5D/3D extension for future work. 
The datasets generated and/or analyzed during the current study are not publicly available, but are available from the corresponding author on reasonable request.

\para{Public benchmarks.}
We evaluate SegDINO on three public medical image segmentation benchmarks.  
TN3K~\cite{gong2023thyroid} is a large-scale thyroid nodule segmentation dataset, containing 3,493 ultrasound images with pixel-level annotations collected from multiple hospitals.  
Kvasir-SEG~\cite{jha2019kvasir} is a polyp segmentation dataset derived from colonoscopy examinations, consisting of 1,000 images with high-quality expert annotations.  
ISIC~\cite{codella2018skin} is a skin lesion segmentation benchmark, providing 2,750 dermoscopic images annotated for lesion boundaries and covering a wide range of lesion types and acquisition conditions. 
\subsection{Implementation Details}
\para{Experimental Settings.} 
For each public dataset, we follow the official training–testing split provided by the organizers to ensure fair comparison.
All images are resized to $256 \times 256$ for consistent input resolution across models, and normalized with the same mean and standard deviation parameters as in DINOv3~\cite{simeoni2025dinov3}. We implement all experiments using the PyTorch framework~\cite{paszke2019pytorch}.
The models are optimized using AdamW~\cite{loshchilov2017decoupled} with a learning rate of $1\times10^{-4}$ and a weight decay of $1\times10^{-4}$.
Cross-entropy loss is employed as the training objective.
Training is conducted for 50 epochs with a batch size of 4.
In this work, we adopt the DINOv3-S backbone, from which intermediate features from the 3rd, 6th, 9th, and 12th layers of DINOv3 are extracted.
All experiments are run on a cloud platform equipped with four NVIDIA RTX A6000 GPUs.

\para{Evaluation Metrics.}
For all datasets, we employ the Dice similarity coefficient (DSC, higher is better) to measure the overlap between predictions and the ground truth, together with the 95th percentile Hausdorff Distance (HD95, lower is better) to evaluate boundary localization accuracy.

\subsection{Comparison with SOTA Methods}

We compare SegDINO with a diverse set of state-of-the-art segmentation models, including U-Net~\cite{ronneberger2015u}, SegNet~\cite{badrinarayanan2017segnet}, R2U-Net~\cite{alom2018recurrent}, Attention U-Net~\cite{oktay2018attention}, TransUNet~\cite{chen2021transunet}, U-NeXt~\cite{valanarasu2022unext}, and U-KAN~\cite{li2025u}.

\para{Quantitative Comparisons.}
Table~\ref{tab:cmp_sota} shows that SegDINO consistently outperforms all competing methods across all four datasets.
On TN3K, SegDINO improves DSC by 3.64\% over the strongest baseline TransUNet, while reducing HD95 by 7.50.
On Kvasir-SEG and ISIC, SegDINO achieves DSC gains of 4.64\% and 2.41\% over SegNet and U-KAN, respectively, accompanied by HD95 reductions of 8.04 and 6.64.
On PanCT, a small lesion segmentation task, SegDINO maintains its advantage by achieving the best DSC and HD95 scores, indicating superior accuracy on small-scale targets.
\begin{table*}[t]
  \centering
  \caption{Comparison with state-of-the-art models.}
  \setlength{\tabcolsep}{1mm}
  \small
  \begin{tabular}{lcccccccc}
    \toprule
    \multirow{2}[4]{*}{Methods} 
      & \multicolumn{2}{c}{TN3K}    
      & \multicolumn{2}{c}{Kvasir-SEG}    
      & \multicolumn{2}{c}{ISIC}
      & \multicolumn{2}{c}{PanCT} \\
\cmidrule{2-9}
      & DSC$\uparrow$ & HD95$\downarrow$
      & DSC$\uparrow$ & HD95$\downarrow$
      & DSC$\uparrow$ & HD95$\downarrow$
      & DSC$\uparrow$ & HD95$\downarrow$ \\
    \midrule
    U-Net~\cite{ronneberger2015u}  & 0.7945 & 24.59 & 0.7916 & 41.58 & 0.8187 & 25.12 & 0.8416 & 5.76 \\
    SegNet~\cite{badrinarayanan2017segnet}  & 0.7924 & 22.74 & 0.8415 & 25.89 & 0.8327 & 21.41 & 0.7966 & 6.97 \\
    R2U-Net~\cite{alom2018recurrent}      & 0.6886 & 27.89 & 0.7367 & 45.64 & 0.8102 & 25.36 & 0.7987 & 5.58 \\
    Att-UNet~\cite{oktay2018attention}     & 0.8015 & 24.64 & 0.8016 & 31.92 & 0.8275 & 26.12 & 0.8373 & 3.36 \\
    TransUNet~\cite{chen2021transunet}    & 0.8027 & 23.95 & 0.8054 & 37.67 & 0.8186 & 24.76 & 0.8465 & 4.93 \\
    U-NeXt~\cite{valanarasu2022unext}       & 0.7285 & 31.40 & 0.6271 & 58.32 & 0.8230 & 23.63 & 0.7885 & 7.82 \\
    U-KAN~\cite{li2025u}        & 0.7866 & 24.49 & 0.7217 & 36.92 & 0.8341 & 23.57 & 0.8111 & 3.82 \\
    \midrule
    \rowcolor{gray!20} \textbf{SegDINO}   
                 & \textbf{0.8391} & \textbf{16.45} 
                 & \textbf{0.8879} & \textbf{17.85} 
                 & \textbf{0.8582} & \textbf{16.93}
                 & \textbf{0.8657} & \textbf{2.61} \\
    \bottomrule
  \end{tabular}
  \label{tab:cmp_sota}
\end{table*}

\begin{figure}[t]
\centering
\includegraphics[width=0.95\columnwidth]{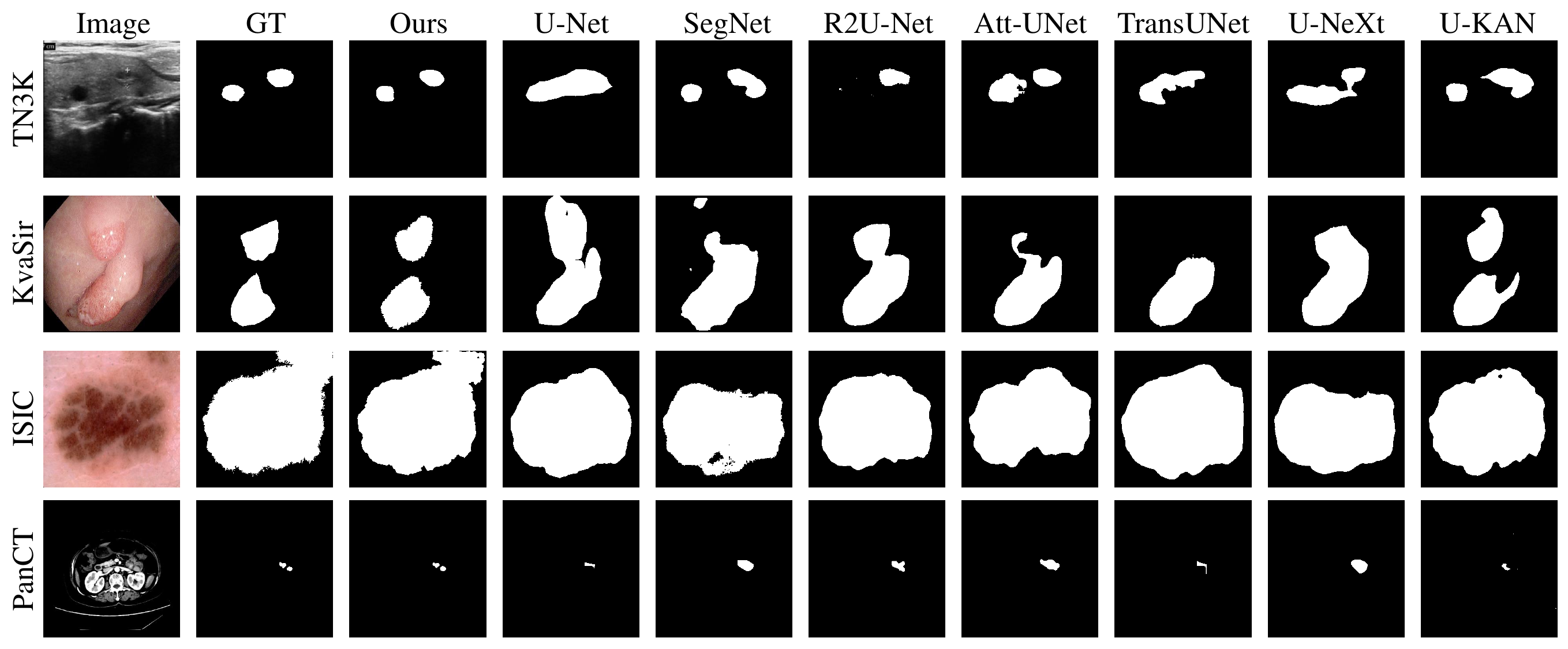}
\caption{
 Visual comparisons of proposed SegDINO and other methods.
} 
\label{fig:vis}
\end{figure}

\para{Visual Comparisons.}
We further present qualitative segmentation results for SegDINO and several competing methods on representative samples from the four datasets.
As shown in Fig.~\ref{fig:vis}, SegDINO produces more accurate object boundaries and cleaner region predictions.
On the PanCT dataset for small-lesion segmentation, SegDINO better captures fine-grained structures.

\begin{figure}[t]
\centering
\begin{subfigure}{0.45\linewidth}
    \centering
    \includegraphics[width=\linewidth]{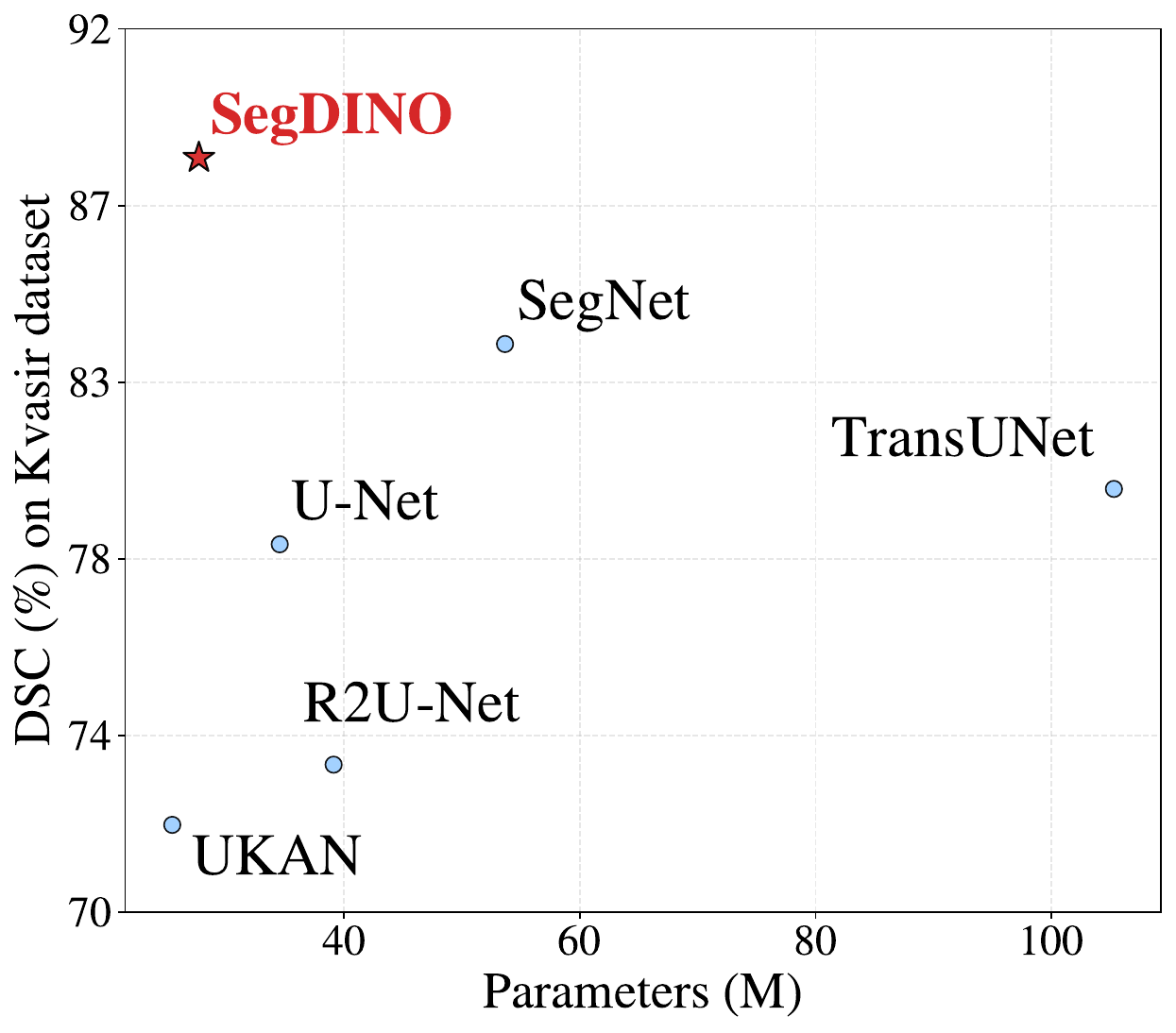}
    \label{fig:params}
\end{subfigure}
\begin{subfigure}{0.45\linewidth}
    \centering
    \includegraphics[width=\linewidth]{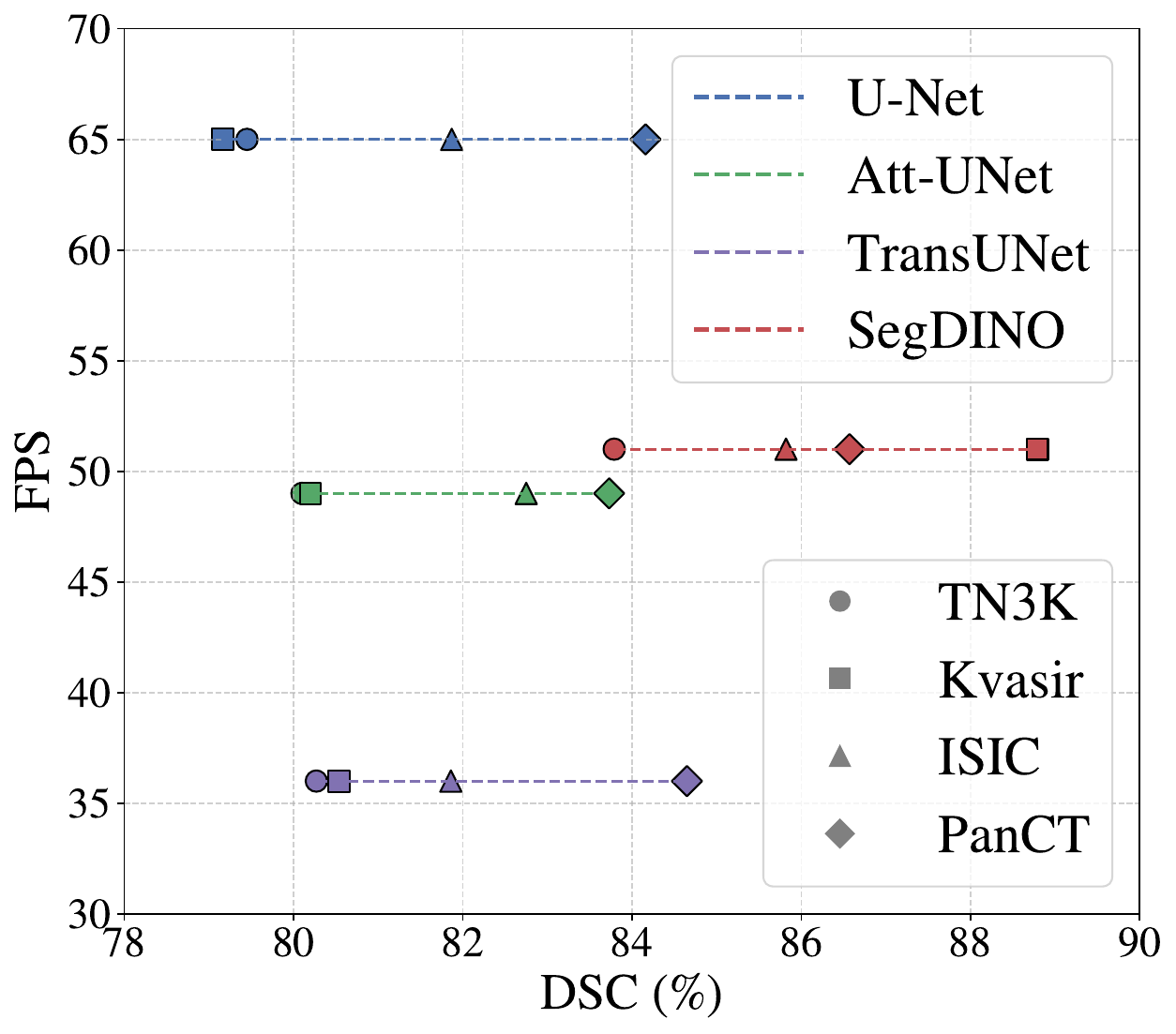}
    \label{fig:fps}
\end{subfigure}
\caption{Overall performance and efficiency comparisons across different datasets.}
\label{fig:fps}
\end{figure}

\para{Efficiency Comparisons.}
As illustrated in Fig.~\ref{fig:fps}, SegDINO achieves strong segmentation performance with low computational overhead.
The model contains 27.68M parameters in total, with 21.60M coming from the DINO-S backbone and 6.08M from the remaining components.
Even with this parameter scale, SegDINO already outperforms most transformer-based counterparts.
Furthermore, SegDINO reaches an inference speed of 51 FPS, surpassing the majority of transformer architectures while remaining competitive with lightweight convolutional models.
These results show that SegDINO maintains a well-balanced combination of accuracy, model size, and inference speed, confirming its effectiveness as an efficient solution for medical image segmentation.

\subsection{Ablation Study}
\begin{table}[t]
\centering
\caption{Ablation study of different modules.}
\small
\setlength{\tabcolsep}{6pt}
\begin{tabular}{l|cc|cc|cc}
\toprule
\multirow{2}{*}{Methods} & \multirow{2}{*}{TPA} & \multirow{2}{*}{SAD} 
& \multicolumn{2}{c|}{TN3K} & \multicolumn{2}{c}{PanCT} \\
\cmidrule(lr){4-5}\cmidrule(lr){6-7}
 &  &  & DSC $\uparrow$ & HD95 $\downarrow$ & DSC $\uparrow$ & HD95 $\downarrow$ \\
\midrule
Basic &  &  & 0.8318 & 18.62 & 0.7758 & 7.97 \\
M1    & \checkmark &  & 0.8379 & 17.79 & 0.8525 & 3.74 \\
M2    &  & \checkmark & 0.8324 & 19.01 & 0.7906 & 7.51 \\
\rowcolor{gray!20} Ours  & \checkmark & \checkmark & \textbf{0.8391} &  \textbf{16.45} & \textbf{0.8657}  & \textbf{2.61} \\
\bottomrule
\end{tabular}
\label{tab:ablation}
\end{table}

We conduct an ablation study to evaluate the contribution of each SegDINO component.
The baseline (\emph{``Basic''}) directly projects multi-level encoder features to a unified channel space and concatenates them at a single resolution for segmentation.
Using TPA alone (\emph{``M1''}) reorganizes encoder features into a pseudo token pyramid.
Using SAD alone (\emph{``M2''}) keeps features at one resolution and performs scale-aware refinement without constructing a pyramid.
The full model combines TPA and SAD.
As shown in Table~\ref{tab:ablation}, TPA consistently boosts performance, and adding SAD further improves results.
Their effects vary by dataset: on TN3K, where targets are large and continuous, both modules yield only marginal gains.
On PanCT, which contains small and fine-grained lesions, TPA brings substantial improvement (e.g., +7.7\% DSC), while SAD alone offers limited benefit.
Overall, TPA is the primary contributor to effective multi-scale representation learning, whereas SAD provides lightweight yet useful refinement.
Given strong DINO-based features, a simple multi-scale hierarchy plus efficient scale-aware decoding is sufficient to exploit the DINO’s inherent information.

\section{Conclusion}
In this work, we presented SegDINO, a lightweight segmentation framework built upon DINOv3 representations. 
Experiments on four datasets show consistent improvements in segmentation accuracy and boundary localization, especially on small-lesion tasks.
Future work will include broader comparisons and ablations, evaluate generalizability across more modalities and clinical scenarios, and explore extensions to 3D volumetric segmentation.
\begin{credits}
\subsubsection{\ackname}
This work was supported by Guangdong Science and Technology Department (2024ZDZX2004) and the National Natural Science Foundation of China (Project No.82572383).
\subsubsection{\discintname}
The authors have no competing interests to declare that are relevant to the content of this article.
\end{credits}
%

\bibliographystyle{splncs04}

\end{document}